\documentclass[10pt,twocolumn,letterpaper]{article}

\usepackage{3dv}
\usepackage{times}
\usepackage{epsfig}
\usepackage{graphicx}
\usepackage{amsmath}
\usepackage{amssymb}
\usepackage{tabu}
\usepackage{multirow}
\usepackage{array}
\usepackage{gensymb}
\usepackage{siunitx}
\usepackage{hyperref}

\newcommand{\norm}[1]{\left\lVert#1\right\rVert}

\usepackage{xcolor}
\usepackage{soul}

\threedvfinalcopy 

\def\threedvPaperID{246} 

\ifthreedvfinal\fi
\begin{document}

\title{3D-Aware Ellipse Prediction for Object-Based Camera Pose Estimation}

\author{Matthieu Zins \qquad Gilles Simon \qquad Marie-Odile Berger\\
Inria, Université de Lorraine, LORIA, CNRS\\
{\tt\small matthieu.zins@inria.fr \tt\small gilles.simon@loria.fr \tt\small marie-odile.berger@inria.fr}

}

\maketitle
\thispagestyle{empty}

\begin{abstract}
In this paper, we propose a method for coarse camera pose computation which is robust to viewing conditions and does not require a detailed model of the scene. This method meets the growing need of easy deployment of robotics or augmented reality applications in any environments, especially those for which no accurate 3D model nor huge amount of ground truth data are available. It exploits the ability of deep learning techniques to reliably detect objects regardless of viewing conditions. Previous works have also shown that abstracting the geometry of a scene of objects by an ellipsoid cloud allows to compute the camera pose accurately enough for various application needs. Though promising, these approaches use the ellipses fitted to the detection bounding boxes as an approximation of the imaged objects. In this paper, we go one step further and propose a learning-based method which detects improved elliptic approximations of objects which are coherent with the 3D ellipsoid in terms of perspective projection. Experiments prove that the accuracy of the computed pose significantly increases thanks to our method and is more robust to the variability of the boundaries of the detection boxes. This is achieved with very little effort in terms of training data acquisition -- a few hundred calibrated images of which only three need manual object annotation. Code and models are released at \url{https://github.com/zinsmatt/3D-Aware-Ellipses-for-Visual-Localization}.
\end{abstract}


\section{Introduction}
In the last few years, deep learning has invested all the fields of computer vision. That of visual localization has not escaped this wave, but the specific nature of this problem has led to  it being approached from a wide variety of angles. On one side,  convolutional neural networks (CNNs) have been trained  to detect visual beacons independently of the camera viewpoint and environmental conditions~\cite{Crivellaro2015,OberwegerECCV2018,TekinSF18}. These methods require an accurate geometric model of the scene  for both training and actual pose computation, e.g. by PnP~\cite{LepetitMF09}.   At the opposite side of the spectrum, the six pose parameters are directly obtained from end-to-end CNN regression, thanks to prior training on data associating images with ground truth poses~\cite{KendallGC15,Kendall2016MUI,Melekhov2017ILU}. Intermediate or alternative approaches have also emerged, making more or less intensive use of geometric models~\cite{Kendall2017GLF,Sundermeyer2018,Xiang-RSS-18}. 

\begin{figure}[tb]
    \centering
    \includegraphics[width=0.94\linewidth]{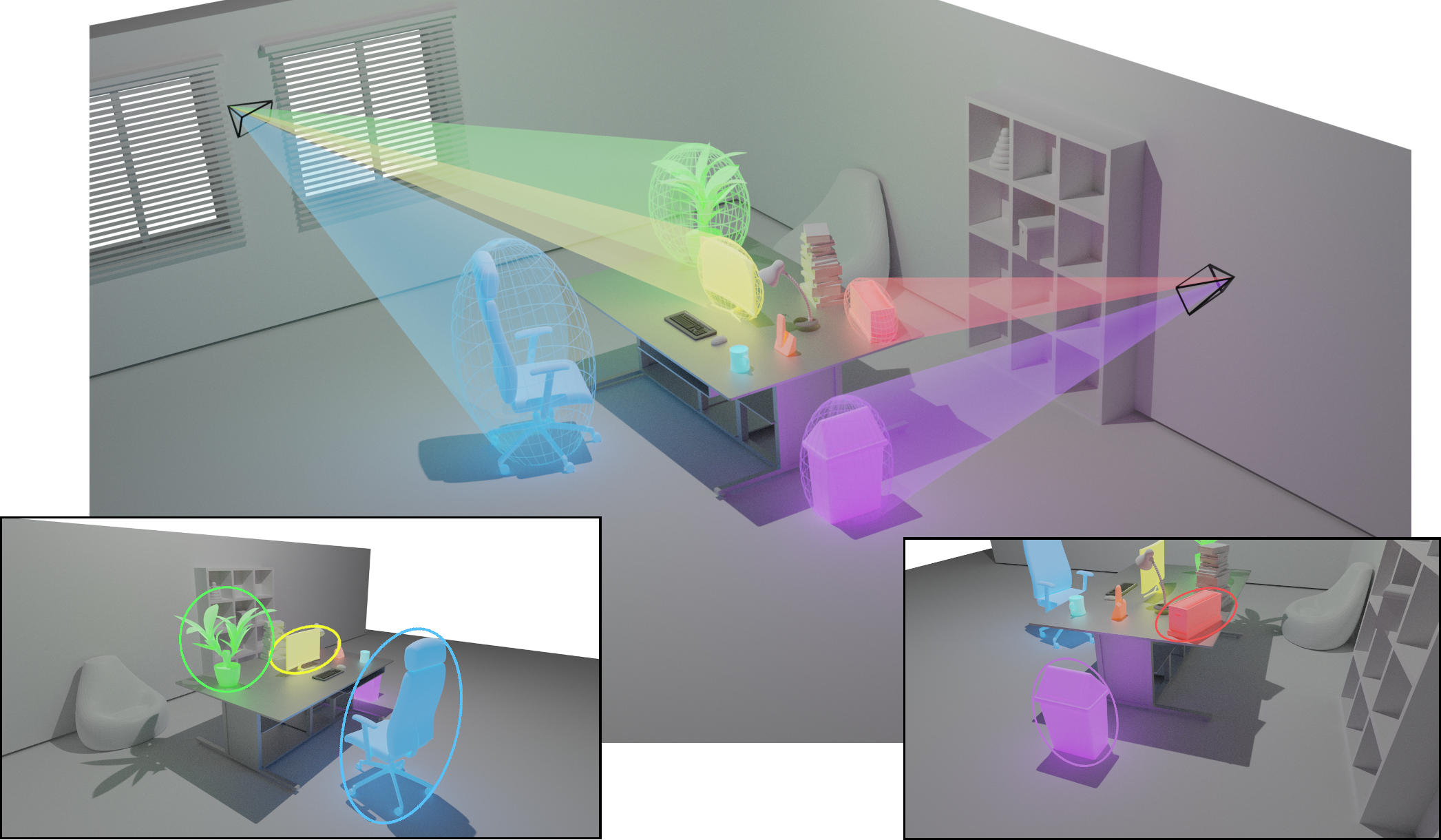}
    \caption{Camera pose estimation from objects.}
    \label{fig:virtual_scene}
\end{figure}

While model-free methods perform  poorly when computing the pose for images distant from the training data~\cite{SattlerZPL19}, obtaining an accurate geometric model of a multi-object scene can be challenging especially when specular or textured-less objects are considered. 

In this paper, we propose a method for coarse pose computation  which  is robust to large changes in  motion and environmental conditions  and  does not require a detailed model of the scene. This method meets the growing need of easy deployment of robotics or augmented reality applications in any environments, especially those  for which no accurate model  nor huge amount of ground truth  data are available.
Our method builds on~\cite{gaudilliere:hal-02170784,gaudilliere:hal-02886633} where  a model-based  method which operates at the level of objects is proposed. Considering objects as features  for pose computation allows us to take advantage of the performance of recognition systems, especially their good invariance to pose and environmental condition changes, while not necessitating systematic retraining of the system when a new scene is considered.
In~\cite{gaudilliere:hal-02170784,gaudilliere:hal-02886633}, pose is computed from   2D/3D object correspondences with 2D objects detected by any object detection algorithm such as YOLO~\cite{RedmonF17}. Approximating 2D objects with ellipses   and 3D objects with ellipsoids  allowed the authors to obtain  direct solutions for pose computation without initial guess (Fig. \ref{fig:virtual_scene}).


Though the method is very promising, the main source of inaccuracy originates from a poor approximation of objects with an ellipse aligned with the image axes and inscribed in the detection bounding box (BB).
In this paper, we go one step further and propose a learning-based  method which detects improved  elliptic  approximations of objects which are coherent  with the 3D ellipsoid i.e. that are likely to be the projection of the ellipsoid. This way of detecting elliptic abstractions of objects significantly improves the accuracy of the recovered pose.
Our main contributions  are as follows:
\begin{itemize}

\item A network for predicting ellipses around objects that are coherent with their 3D ellipsoidal abstractions. Its goal is to overcome the weaknesses of directly fitting the ellipses to the axis-aligned BBs. We propose a data augmentation procedure that allows for robustness to box boundaries variability. 
We also show that, even if the box boundaries contain a part of background that might be learnt by the network, our method is almost unaffected in typical scenarios, where the scene does not change between training and testing.


\item
We show how the concept of ellipsoidal abstractions of objects and 3D-coherent ellipse predictions can be used for robust pose computation when only a small amount of data is available on the scene. We show that the  pose accuracy  little depends on the choice of this ellipsoidal abstraction, which makes the method flexible and easy to use in practice. Only three calibrated images need to be annotated by hand to  build  the ellipsoid cloud. Annotations of the object are then obtained  by projection in  the training images.
\end{itemize}
Experiments  prove  that the accuracy of the computed pose dramatically increases thanks to our 3D-aware ellipse detection procedure  and is close to the accuracy obtained with  learning-based methods that require an accurate 3D model for  training.

\section{Background and related works}

Pose computation from monocular RGB images is an important problem in computer vision which witnessed a complete renaissance with the emergence of deep learning. Thanks to the ability of such methods to detect features across a wide range of viewpoints, largely independently from environmental conditions, this opens the way towards more robust localization and matching methods, especially able to handle few-textured scenes.

One of the pioneering method in the use of deep learning for pose computation is poseNet~\cite{KendallGC15} where absolute camera pose is regressed. Besides the difficulty of balancing rotation and translation errors in the cost function, a major default of such approaches, pointed out by Sattler~\cite{SattlerZPL19}, is that direct pose regression is more closely related to pose approximation via image retrieval than to accurate pose estimation via 3D structure. As a result, such methods generalize poorly to trajectories far from the training sequences. Other methods propose to regress dense 3D scene coordinates~\cite{BrachmannR18, BuiAIN18} and estimate the camera pose by solving a PnP problem. These methods were coupled with advanced versions of RANSAC~\cite{BrachmannDSAC} and obtain very good performance, but only on small-scale scenes.

In order to scale to large environments, \cite{Weinzaepfel_2019_CVPR} propose a method that relies on the detection on a set of objects of interest and exploits dense 2D/3D correspondences between them and reference images to estimate the pose. However, their method is limited to planar objects.

Finding the pose of the camera from general shape objects can also be viewed as estimating the objects pose in the camera frame. Many works exist on this subjects \cite{kehl2017, CDPN, PVNet,  RadL17, Sundermeyer2018, TekinSF18, DPOD}. They usually assume to have access to a detailed model of the object in order to generate synthetic training data and, thereby enlarge the 6D pose space where pose can be correctly recovered. SSD-6D~\cite{kehl2017} extends the idea of 2D object detection and infers 6D pose based on a discrete viewpoint classification while an autoencoder is used in~\cite{Sundermeyer2018} to recover the object orientation. Another way to infer object pose is by predicting the 2D projections of the corners of the bounding box of the 3D object with a CNN. This avoids the need for a meta-parameter to balance the position and orientation error since the 6D pose can be estimated with PnP from 2D/3D correspondences. In BB8~\cite{RadL17}, segmentation is first performed to detect the objects and a CNN then infers the projection of the BBs. Data augmentation with a random background is performed during training to reduce the influence of the scene context. Building on YOLO and BB8, Tekin \etal~\cite{TekinSF18} proposed a single-shot deep CNN architecture which directly predicts the 2D projections of the 3D BB vertices in the image.

The BB of detected objects can be also used for localization issues with the idea of decoupling the rotation and the translation. In PoseRBPF~\cite{deng2019}, and PoseCNN~\cite{Xiang-RSS-18} the translation is constrained to be along the vector from the camera center to the center of the BB. $T_z$ is then approximately determined from the ratio between 3D ROI and 2D BB sizes. An auto-encoder trained on the synthetic rendering of the 3D object is used for rotation estimation in PoseRBF, while PoseCNN regresses a quaternion for rotation and a mask, both with the need of an accurate model of the object.
Without accurate knowledge of the objects, but under some assumptions on the scene, e.g. objects lying on a flat scene, Li \etal~\cite{Li2017} used object detections to estimate relative camera poses for viewpoints changes. Wang \etal~\cite{NOCS} proposed a category-level approach to predict normalized objects coordinates without the need of detailed 3D models. However, the method is not straightforward for new objects from unseen categories and depth information is needed to compute the pose. In the context of autonomous driving,~\cite{MousavianAFK2017} estimates the pose and the dimensions of an object’s 3D BB from a 2D BB and the surrounding image pixels under additional constraints on the box orientation. However, modelling objects as 3D cuboids and the 2D detections by rectangles does not allow to derive closed-form solutions to projection equations and leads to solutions with a high combinatorics. 
 

Modeling 2D/3D objects correspondences based on ellipses/ellipsoids was already used by~\cite{rubino2018} in the context of multiview reconstruction and by Nicholson \etal in~\cite{nicholson2019} in the context of SLAM. Resolution was based on the minimization of a geometric cost function over the six camera parameters, using odometry sensors for initial position and orientation. Recent works have proposed direct solutions without the need of initial prior for pose computation from ellipse-ellipsoid matches:~\cite{gaudilliere:hal-02170784} show that the problem of estimating the camera pose from ellipse-ellipsoid correspondences has at most 3 degrees of freedom, since the position can be obtained from its orientation. Direct closed form solution can thus be estimated once the orientation is known. In ~\cite{gaudilliere:hal-02886633}, a direct method for full recovery of the pose from more than 2 ellipse-ellipsoid correspondences was proposed under assumptions satisfied by many robotics applications. In practical experiments, axis-aligned ellipses are inferred from the BBs detected by YOLO. The authors however note that such an elliptic 2D approximation is not always sufficiently accurate and may lead to a significant error on the estimated pose.
 
Instead of just detecting a rectangular BB of 2D objects, it is also possible to detect elliptic objects. An immediate idea is to perform instance segmentation, as in Mask R-CNN~\cite{maskRCNN} and directly fit an ellipse to the output mask. A recent work~\cite{dong2020ellipse} shows that such a strategy fails to capture the ellipse orientation and proposes a CNN for ellipse regression. Such a work is typically dedicated to the detection of objects with elliptic shapes. Besides the fact that the detection is not aware of a particular 3D model, another difference with our work is that we are interested in detecting ellipsoids which approximate an object, whose shape may be quite different from a perfect ellipsoid.

\section{System overview}

\begin{figure}[t]
    \centering
    \includegraphics[width=\linewidth]{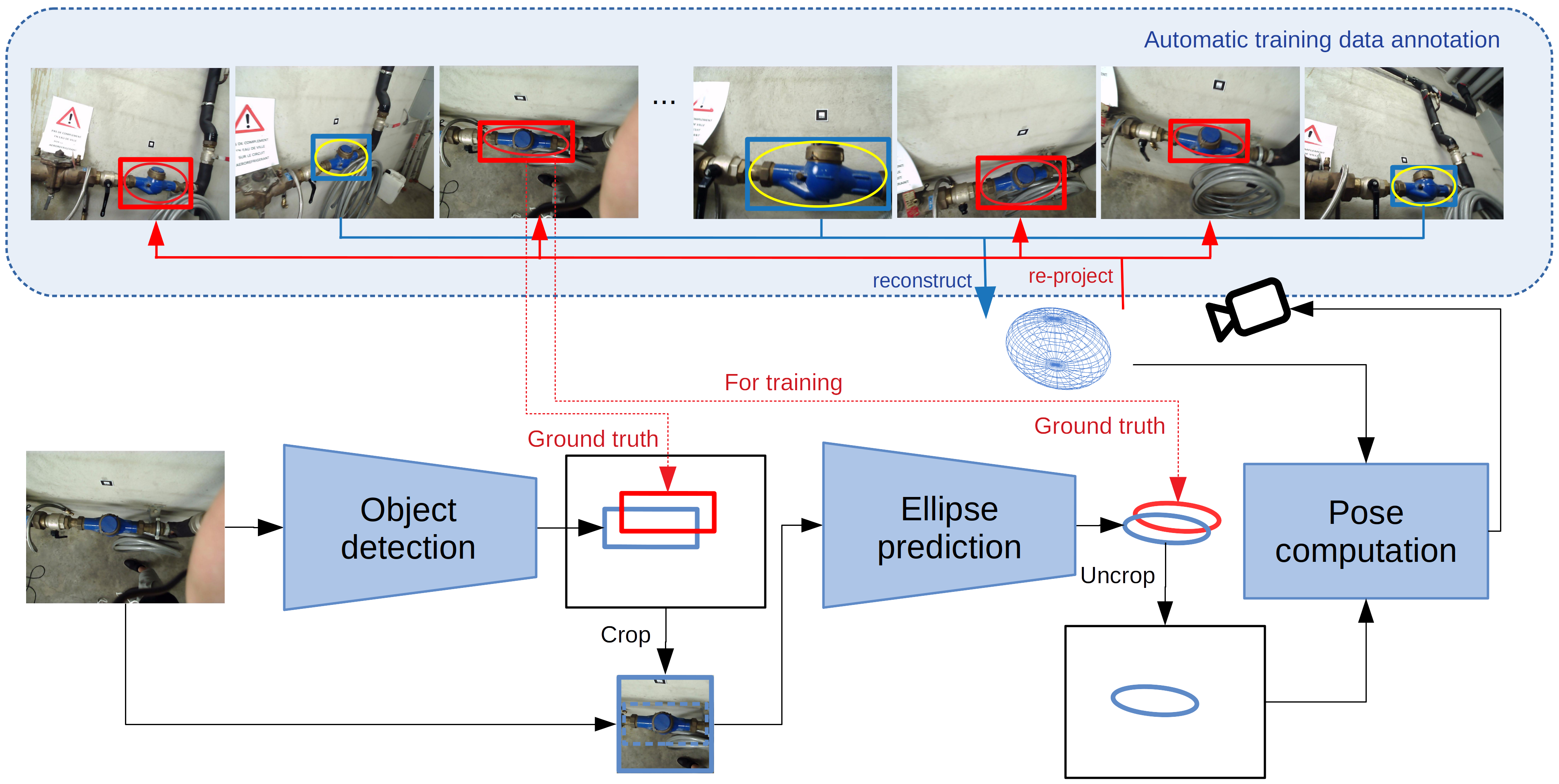}
    \caption{Overview of our full system. The new ellipse prediction block is the key part of our work.}
    \label{fig:system}
\end{figure}

Our method uses an ellipsoid cloud as a light scene model, where each ellipsoid approximates an object. The pose computation system consists of three main components (Fig.~\ref{fig:system}, bottom part). The first component is a CNN designed to detect objects as BBs. The second component is another CNN aimed at predicting, for each BB, the ellipse corresponding to the projection of the object's ellipsoid according to the camera pose. Unlike global approaches, such as PoseNet, which apply to the whole image, our prediction network uses local patches around each detected object. This has the advantage of restricting a lot the change of appearance and should enable our prediction network to better interpolate new views. Finally, the last component aims at calculating the pose from the ellipses thus obtained, the ellipsoid cloud and the labels associated with these features. 

The object detection network takes as input an image and returns the BBs of the detected objects, with their class.  Many networks have already been proposed to solve this task very efficiently~\cite{RedmonF17,FasterRCNN} and many types of objects are already recognized by these networks. It is, moreover, easy to fine-tune a pre-trained network in order to make it recognize new objects, as only a few images are required for that purpose.
In order to preserve  the aspect ratio of the detected object, the crop box is completed  with the content of the image to meet  the ellipse prediction network's input requirement. From these cropped images, the network predicts the ellipse parameters expressed in the BB frame. All the ellipses are then transferred to the initial image frame according to the coordinates of the BBs. The transferred ellipses, associated with the detection labels, are then used along with the ellipsoid cloud to compute the pose. 

Pose computation from one or more ellipse-ellipsoid pairs is based on the work described in~\cite{gaudilliere:hal-02170784,gaudilliere:hal-02886633}. The method used depends on whether or not external orientation data are available. Measuring  camera orientation from image content or sensor acquisitions is generally easier to do than estimating the position. Indeed, position sensors are often inaccurate (GPS) or have a limited range and are sensitive to environmental disturbances (magnetic, acoustic sensors, etc.). By contrast, automatic vanishing point detection may lead to accurate orientation measures as long as the scene satisfies the Manhattan world assumption~\cite{gaudilliere:hal-02170784}. On the other hand, IMU sensors can be used, limiting drift and expressing orientations in an absolute reference frame by taking into account the Earth's magnetic field and gravity. A very interesting result from a practical point of view is that knowing the orientation of the camera, a closed form solution makes it possible to deduce its position from a single ellipse-ellipsoid pair~\cite{gaudilliere:hal-02170784}. 
If the orientation is available, we therefore use a RANSAC procedure with minimal sets of one ellipse-ellipsoid pair~\cite{gaudilliere:hal-02170784}. Multiple association hypotheses may occur in case objects of the same class appear in the image, but this only adds as many correspondences to be included in the RANSAC draws, preserving a linear algorithmic complexity. If orientation data is not available, two  ellipse-ellipsoid pairs are at least required to compute the pose. We then use RANSAC with minimal sets of two correspondences~\cite{gaudilliere:hal-02886633}. 



\section{Ellipse prediction}
\subsection{Motivation for 3D supervision}
\label{ssec:intuition_for_3d}

\begin{figure}
    \centering
    \includegraphics[width=0.61\linewidth]{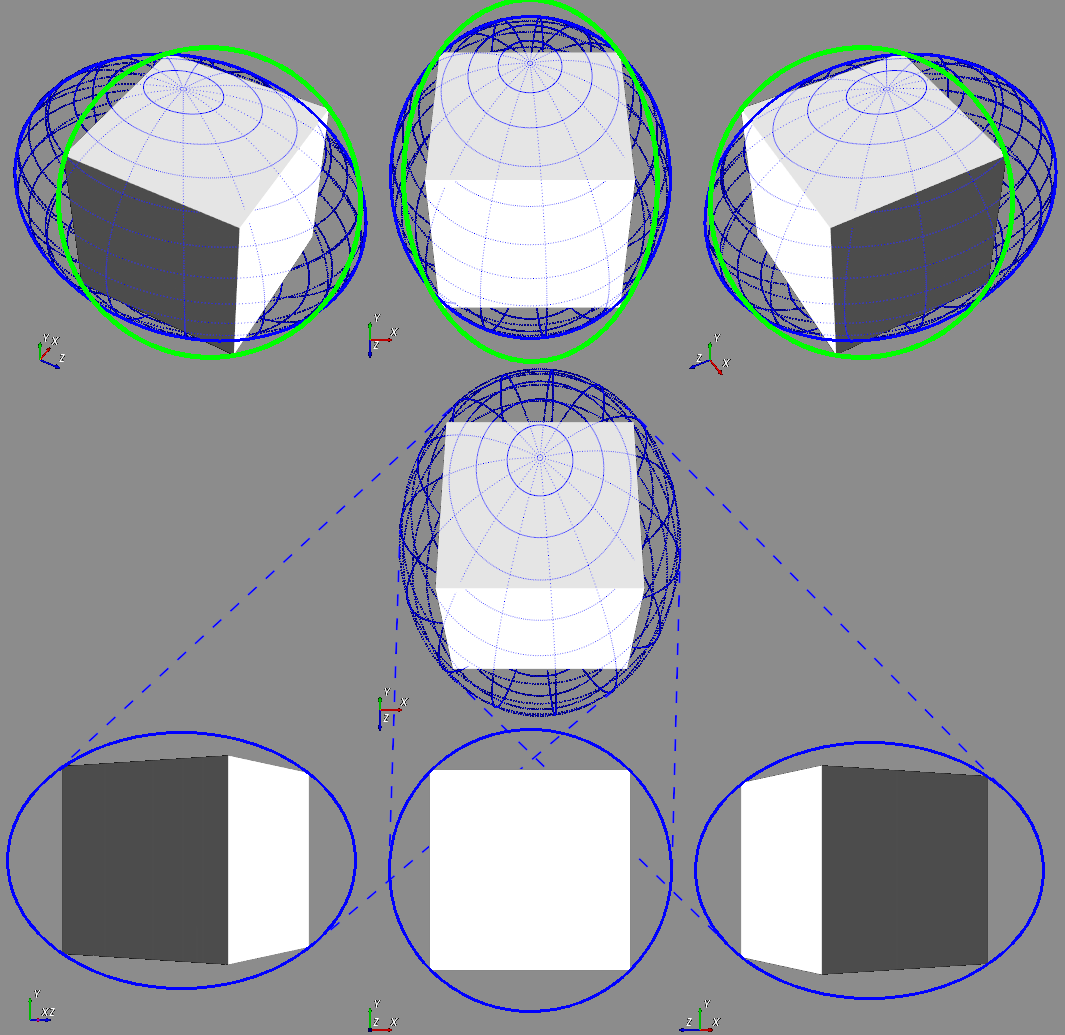}
    \caption{Bottom: Images and detections used for reconstruction. Middle: Reconstructed ellipsoid. Top: Projection in new views.}
    \label{fig:synthetic_2D_only}
\end{figure}

We give here an intuition of why a 2D-only supervision is not sufficient for detecting good ellipses. Pose estimation from ellipse-ellipsoid pairs comes down to aligning cones, which means that the accuracy is heavily influenced by how much the ellipse detection satisfies the real projection of the ellipsoid. In Fig.~\ref{fig:synthetic_2D_only}, we simulated a "perfect" 2D detector by finding the minimum-area ellipse containing the object and reconstructed an ellipsoid from three views. We can  observe that the 2D detections in other views are not fully coherent with the projection of this ellipsoid. In fact, the deformations caused by perspective, the viewpoints used for reconstruction and the shape of the object have a significant impact. This is why our method relies on 3D supervision and manages to be almost unaffected by these factors.

\subsection{Ellipse parameterization}

Different parameterizations of an ellipse can be used, among which the following were considered:
(i) three points (center and  two points corresponding to the axes ends),
(ii) five parameters of the quadratic form matrix,
(iii) center, axes lengths and orientation angle.

Using the five parameters of the quadratic form  gives a minimal representation, because an ellipse has five degrees of freedom. However, these parameters have no physical interpretation which makes the associated L2 distance loss hard to optimize.
Parameterizing an ellipse by three points has the advantage of having a real physical sense. The first point is chosen at the center of the ellipse and the  other two points at the end of each axis. An inconvenient of this representation is that there is no implicit constraint to force orthogonality. Instead, we could choose to represent the ellipse with its center,  axes lengths and  orientation angle. Both parameterizations have ambiguities due to the ellipse symmetries. A canonical form can be obtained by enforcing the first axis to be the largest one, its endpoint to be in the right half-circle and the second axis endpoint to be in the bottom half-circle. Also, the angle is measured between the horizontal axis and the largest axis. This parameterization, however, implies discontinuities in the orientation angle when the ellipse is almost vertical. Moreover, L2 loss functions with non-normalized heterogeneous quantities are hard to optimize (e.g. the camera position and orientation parameters in PoseNet~\cite{KendallGC15}) are hard to optimize.

To tackle these issues, we use the center, axes and orientation parameterization but with a classified, rather than regressed, orientation angle~\cite{MousavianAFK2017}. More specifically, this angle is discretized  into \textit{n} overlapping bins distributed between $-\frac{\pi}{2}$ and $\frac{\pi}{2}$. 
Our network predicts the corresponding bin but also regresses a correction angle as a value added to the mid-angle of the predicted bin.

\subsection{MultiBin Loss}

The center and dimensions of the ellipse are both optimized using a L2 loss, respectively
\begin{align}
    \mathcal{L}_{center} &= \frac{1}{N} \norm{c_{gt} - c_{pred}}^2\\
     \mathcal{L}_{dim} &= \frac{1}{N} \norm{d_{gt} - d_{pred}}^2
    \end{align}



where $c_{gt}$ and $c_{pred}$ are the ground truth and predicted centers, $d_{gt}$ and $d_{pred}$ are the ground truth and predicted dimensions, and $N$ is the number of images in the batch.

As proposed in~\cite{MousavianAFK2017}, the angular part has been split into both classification and regression. We use a standard \textit{cross-entropy} loss for the classification part ($\mathcal{L}_{bin}$) and a correction loss is added to reduce the difference between the predicted angle and the ground truth angle in each of the bins that overlaps with this angle (one or two bins). 

\begin{equation}
    \mathcal{L}_{correction} = -\frac{1}{n_{\theta^*}} \sum_i^{n} o_i \cdot cos(\theta^* - \sigma_i - \Delta \theta_i)
\end{equation}
where $\theta^*$ is the ground truth angle, $n_{\theta^*}$ is the number of bins that overlap the ground truth angle (1 or 2), $o_i$ is either 0 or 1 and indicates if bin $i$ overlaps the ground truth angle, $\sigma_i$ is the mid-angle of the i-th bin and $\Delta \theta_i$ is the predicted correction angle that is applied to the mid-angle of the bin. This loss has effects only on the correction angles of the bins that actually overlap with the ground truth angle.

The final loss of our network is a weighted combination of each individual loss.
\begin{equation}
    \mathcal{L} = \alpha (\mathcal{L}_{center} + \mathcal{L}_{dim}) + (\mathcal{L}_{bin} + \beta \mathcal{L}_{correction})
\end{equation}

The parameters $\alpha$ and $\beta$ are used to balance the loss intensities and were determined empirically. We used the same values for all experiments (respectively $0.01$ and $1.0$).

\subsection{Network architecture}
\begin{figure}
    \centering
    \includegraphics[width=\linewidth]{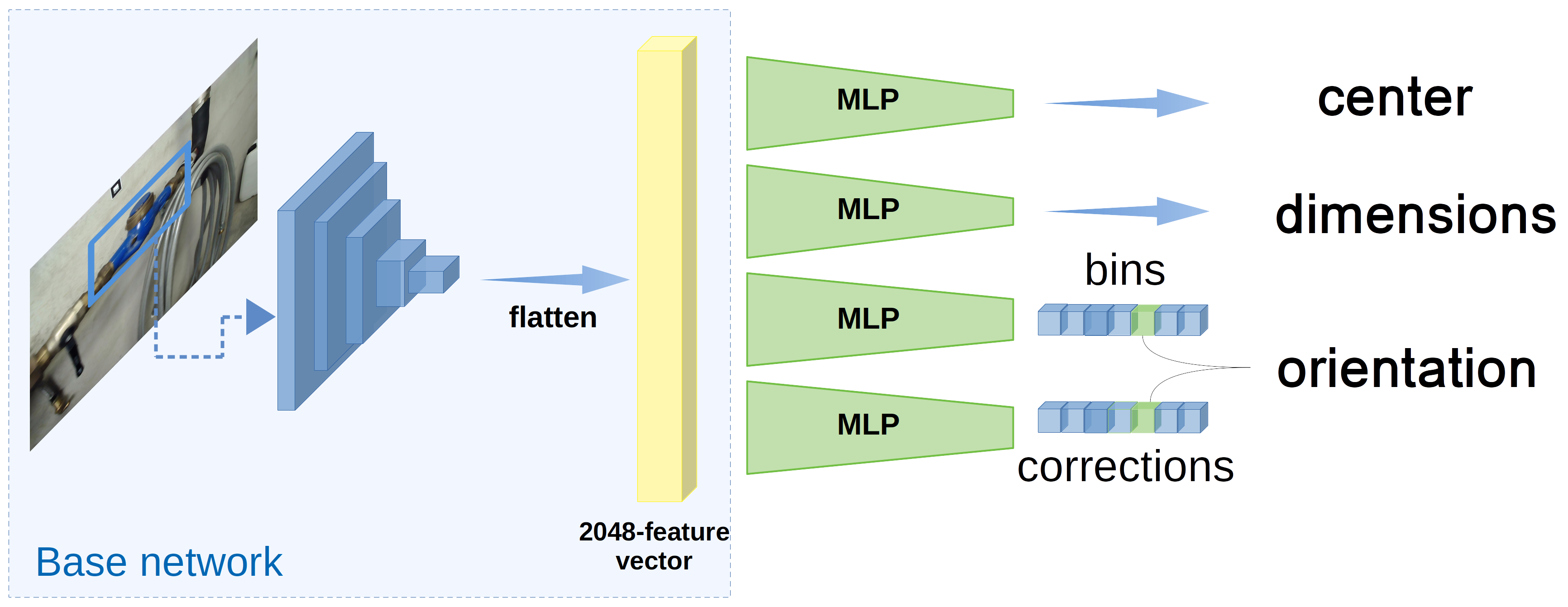}
    \caption{Ellipse prediction network.}
    \label{fig:architecture}
\end{figure}

Our architecture (Fig.~\ref{fig:architecture}) includes a base convolutional network used for features extraction, followed by four parallel branches.  We used a pre-trained VGG-19 network as base and just cut off the final \textit{softmax} part, which is traditionnally used for classification. Then, the parallel branches take these features as input and consist of three-layers fully connected networks with \textit{ReLU}  as non-linear activations.

The \textit{center} branch predicts two values corresponding to the center coordinates. The \textit{dimensions} branch regresses the width and height of the ellipse. The \textit{bin} branch predicts $n$ values, where $n$ is the number of bins. These values correspond to scores or confidences of belonging to a specific bin and are then passed to a softmax layer. Finally, the \textit{correction} branch regresses $2n$ values,  $cos(\Delta \theta_i)$ and $sin(\Delta \theta_i)$ for each bin. To get valid cosine and sine values, a normalization step was added at the end of this branch.

\subsection{Network architecture validation}
\begin{figure}
    \centering
    \includegraphics[width=0.9\linewidth]{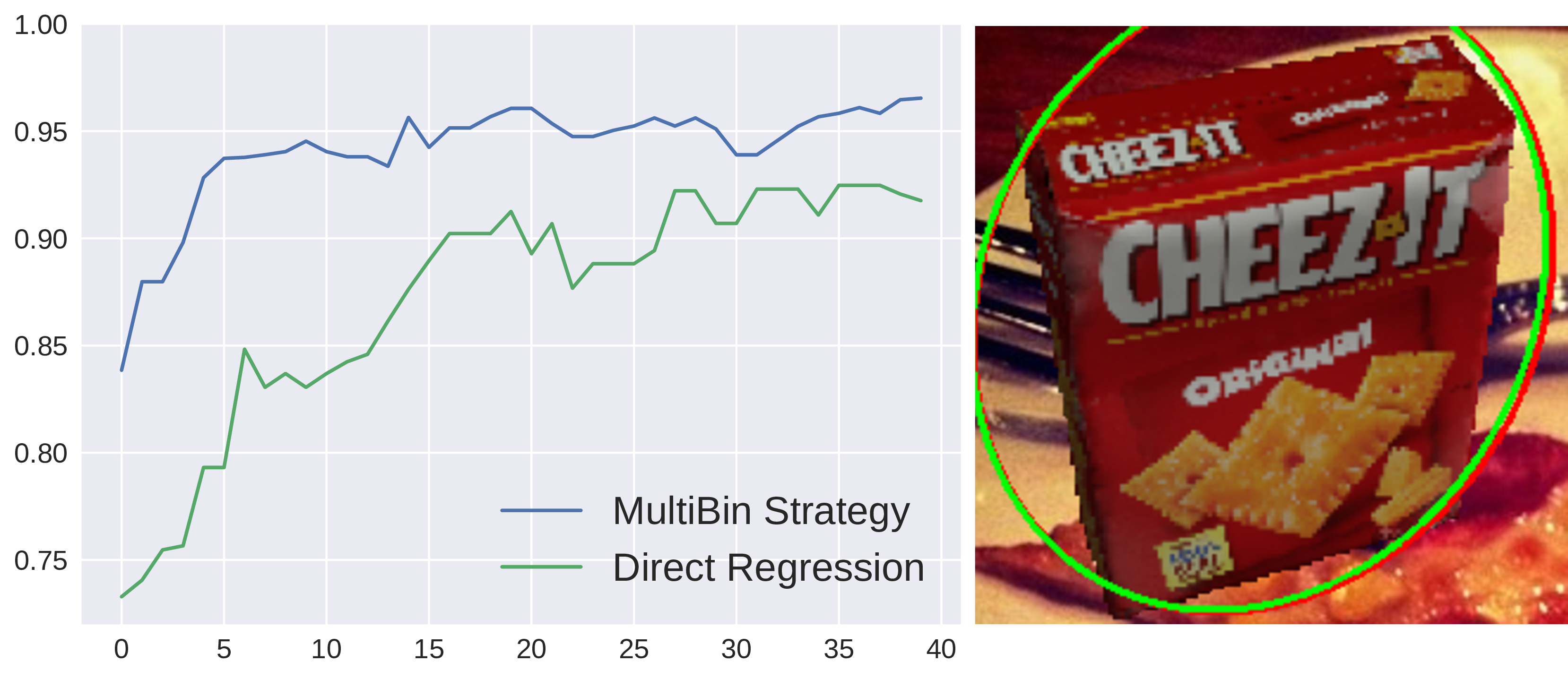}
    \caption{Left: Mean IoUs obtained on test images for an increasing number of training epochs. Right: Example of predicted ellipse (green). The ground truth ellipse is in red.}
    \label{fig:network_validation}
\end{figure}

We validated our network architecture and loss function on synthetic data, consisting of multiple views of an object with the camera placed on a semi-sphere around it and random background from COCO dataset~\cite{COCO}. We sampled 100 values of azimuth and 20 of elevation for training and test data were generated at intermediate positions.
For evaluation, we computed the \textit{IoU} between the predicted ellipse and the ground truth one. We obtained very high IoUs (more than 0.95) on the test images, keeping in mind that the viewpoints coverage of the training was quite large.
 Figure \ref{fig:network_validation} also shows the advantage of the multibin strategy over the direct regression.

\section{Data acquisition and augmentation}
\label{sec:data}

The only prerequisite for our system is that images of the scene have been acquired and calibrated from various viewpoints, i.e. we know for each training image the intrinsic parameters of the camera as well as its position and orientation in a global frame. This is the  minimum   prerequisite  of any visual pose computation methods using deep learning. In practice, relatively few images are needed  (around 200) both to fine-tune the detection network (if required) and to train the prediction network. Obtaining these data  can be done in several ways, e.g. by using a differential GPS coupled with orientation sensors,  a Structure-from-Motion (SFM) technique, or a marker \cite{yang:hal-02735272}.  





\subsection{3D model and training data generation}
\label{ssec:data_generation}


Once the calibrated images  have been obtained, we propose a  practical method to get both the ellipsoid cloud and the  ground truth boxes and ellipses in all training images   by just labeling a few boxes in a small number of them. 
It consists of the following steps (Fig.~\ref{fig:system}, top):
    (i) choose a minimum of three images of the scene showing the object(s) from various viewing angles, 
    (ii) define boxes around the objects visible in these images (blue boxes  Fig.~\ref{fig:system}, top) and associate a label to each box,
    (iii) fit ellipses to these boxes, with axes parallel to those of the boxes, 
    (iv) build the ellipsoid cloud from the 
    2D ellipses and projection matrices; an analytic solution was obtained by  Rubino {\it et al.} to solve this problem from only three calibrated views~\cite{rubino2018},
    (v) reproject the ellipsoid cloud in all training images 
    by using the ground truth projection matrices. The obtained ellipses (in red in Fig.~\ref{fig:system}) will be used as ground truth data for training the ellipse prediction CNN.
    (vi) If needed, the BBs of the projected ellipses  (red boxes in Fig.~\ref{fig:system}) will be used as ground truth BBs to fine-tune the object detection CNN.
    

The ellipsoids obviously depend  on the  images chosen for reconstruction. Fortunately, we show in Section~\ref{sec:experiments} that their size and orientation  may vary significantly without degrading the method performance. This is due to the fact that the ellipsoids are only intermediate data used to calculate the pose from the detected ellipses, themselves predicted to match the perspective projection of the  ellipsoids. The only critical point is that  the ellipsoids used to generate the training data must be those used at runtime to compute the pose. Also, the internal calibration of the camera does not change between training and testing.


\subsection{Data augmentation}
\label{ssec:data_augmentation}
Data augmentation plays an important role in the training of our prediction network and its generalization with a relatively limited number of annotated images. Several strategies were performed online (during training):
(i) Color jittering randomly changes the brightness, contrast and saturation of an image in order to simulate illumination changes.
(ii) Blurring filters the images with a randomly-sized Gaussian kernel in order to accommodate different resolutions caused by the object distance.
(iii) Shifting randomly translates the images so that the object is not always perfectly centered, which should accommodate noisy object crops.
(iv) In-plane rotations as well as (v) perspective deformations (homographies) were added to generate new views of the object. The last two can, for example, simulate a camera which is not held upright, or not aiming at the object center.

\section{Experiments}
\label{sec:experiments}



\subsection{Metrics}

In our experiments, we used three common metrics: reprojection error (in pixels), ADD (as in~\cite{HinterstoisserLIHBKN12, kehl2017, RadL17}) and camera pose error (either only the position or the full 6D pose). Unless otherwise specified, the percentages in the tables  represent the proportion of test images that satisfy a certain threshold.

\subsection{Camera position estimation}
We evaluated the accuracy of the camera position estimated using our predicted ellipses on the LINEMOD dataset~\cite{HinterstoisserLIHBKN12}, which provides RGBD images of 15 objects in cluttered environments with ground truth pose. Note that we did not perform specific retraining for object detection in LINEMOD and T-LESS and used the 2D boxes given by the reprojected 3D object.

\begin{figure}
    \centering
    \includegraphics[width=\linewidth]{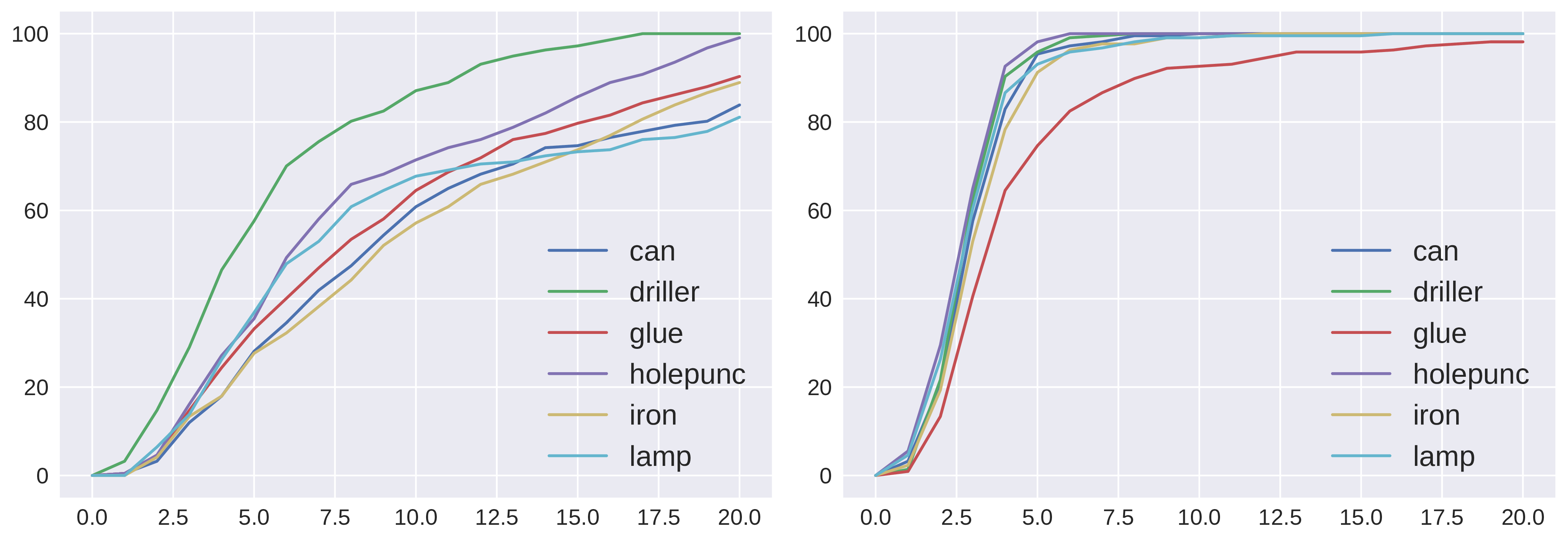}
    \caption{Proportion of positions correctly estimated  wrt. the error threshold (in cm) on the horizontal axis. Left: using the inscribed ellipse. Right: using the predicted ellipse.}
    \label{fig:Linemod_position_error}
\end{figure}

Using an initial ellipsoid, obtained by reconstruction from a few images, we generated ellipse annotations and trained our prediction network. We limit this experiment to the camera position because only one object can be used per sequence (the rest of the scene changes). Thus, the camera orientation is assumed to be known. To be realistic, we added a random noise uniformly sampled in [\SI{-2}{\degree}, \SI{2}{\degree}] on each of its Euler angle at test time. 
Figure~\ref{fig:Linemod_position_error} and Tables~\ref{tab:Linemod_reproj} and~\ref{tab:Linemod_ADD} show the results obtained in terms of reprojection error, position error and ADD. The experiments show a significant improvement of the position estimation, compared to Gaudilliere \etal \cite{gaudilliere:hal-02170784} (inscribed ellipse) and the obtained accuracies are comparable with Tekin \etal \cite{TekinSF18}, a state-of-the-art method trained using a detailed model of the object.

\begin{table}[t]
  \scriptsize%
	\centering%
    \begin{tabu}{@{\hspace{0mm}}|@{\hspace{2mm}}c@{\hspace{2mm}}|@{\hspace{2mm}}c@{\hspace{2mm}}|@{\hspace{2mm}}c@{\hspace{2mm}}c@{\hspace{2mm}}c@{\hspace{2mm}}c@{\hspace{2mm}}|@{\hspace{2mm}}c@{\hspace{2mm}}c@{\hspace{2mm}}c@{\hspace{2mm}}|@{\hspace{0mm}}}
  \hline
 
    &  Tekin & \multicolumn{4}{@{\hspace{2mm}}c@{\hspace{2mm}}|@{\hspace{2mm}}}{Gaudilliere} & \multicolumn{3}{c|}{} \\
   \textbf{Method} &  \etal & \multicolumn{4}{@{\hspace{2mm}}c@{\hspace{2mm}}|@{\hspace{2mm}}}{\textit{et. al.}} & \multicolumn{3}{c|}{Ours}  \\
    & \cite{TekinSF18} & \multicolumn{4}{@{\hspace{2mm}}c@{\hspace{2mm}}|@{\hspace{2mm}}}{\cite{gaudilliere:hal-02170784}} &  \multicolumn{3}{c|}{} \\
   \hline
   \textbf{Thresh.} & 5 px & 5 px & 10 px & 15 px & 20 px & 5 px & 10 px & 15 px\\
  \hline
ape     & 92.10 & 95.39 & 100.0 & 100.0 & 100.0 & \textbf{100.0} & 100.0 & 100.0  \\
cam     & 93.24 & 49.77 & 94.47 & 100.0 & 100.0 & \textbf{94.47} & 100.0 & 100.0  \\
can     & 97.44 & 57.60 & 79.26 & 98.62 & 100.0 & \textbf{99.54} & 100.0 & 100.0  \\
cat     & \textbf{97.41} & 68.20 & 98.62 & 100.0 & 100.0 & 90.32 & 100.0 & 100.0 \\
driller & 79.41 & 16.13 & 61.75 & 90.32 & 98.62 & \textbf{96.77} & 100.0 & 100.0  \\
duck    & 94.65 & 89.40 & 100.0 & 100.0 & 100.0 & \textbf{100.0} & 100.0 & 100.0  \\
eggbox  & 90.33 & 97.70 & 100.0 & 100.0 & 100.0 & \textbf{100.0} & 100.0 & 100.0 \\
glue    & \textbf{96.53} & 54.38 & 88.02 & 95.85 & 99.54 & 94.93 & 99.08 & 100.0 \\
holepunc& 92.86 & 83.41 & 100.0 & 100.0 & 100.0 & \textbf{99.54} & 100.0 & 100.0 \\
iron    & 82.94 & 17.05 & 51.15 & 78.34 & 93.09 & \textbf{90.78} & 99.54 & 100.0  \\
lamp    & 76.87 & 18.43 & 60.37 & 84.79 & 97.24 & \textbf{92.63} & 99.54 & 100.0 \\
phone   & 86.07 & 34.56 & 70.97 & 88.48 & 97.24 & \textbf{90.78} & 100.0 & 100.0 \\

  \hline
  \end{tabu}%
  \caption{Reprojection error.}
    \label{tab:Linemod_reproj}
\end{table}

\begin{table}[t]
  \scriptsize%
	\centering%
  \begin{tabu}{@{\hspace{0mm}}|@{\hspace{2mm}}c@{\hspace{2mm}}|@{\hspace{2mm}}c@{\hspace{2mm}}|@{\hspace{2mm}}c@{\hspace{2mm}}c@{\hspace{2mm}}c@{\hspace{2mm}}|@{\hspace{2mm}}c@{\hspace{2mm}}c@{\hspace{2mm}}c@{\hspace{2mm}}|@{\hspace{0mm}}}
  \hline
 
    & Tekin & \multicolumn{3}{@{\hspace{2mm}}c@{\hspace{2mm}}|@{\hspace{2mm}}}{Gaudilliere} & \multicolumn{3}{c|}{} \\
   \textbf{Method} & \etal & \multicolumn{3}{@{\hspace{2mm}}c@{\hspace{2mm}}|@{\hspace{2mm}}}{\etal} & \multicolumn{3}{c|}{Ours}  \\
    & \cite{TekinSF18} & \multicolumn{3}{@{\hspace{2mm}}c@{\hspace{2mm}}|@{\hspace{2mm}}}{\cite{gaudilliere:hal-02170784}} &  \multicolumn{3}{c|}{} \\
   \hline
    \textbf{Threshold} & & \multicolumn{3}{@{\hspace{2mm}}c@{\hspace{2mm}}|@{\hspace{2mm}}}{} & \multicolumn{3}{@{\hspace{2mm}}c@{\hspace{2mm}}|@{\hspace{0mm}}}{}\\

   \textbf{(\% of diam.)} & 10\% & 10\% & 15\% & 25\% & 10\% & 15\% & 25\%\\
  \hline

      ape & 21.62 &	18.43 &	35.94 &	56.68 &	\textbf{64.98} &	80.65 &	93.09 \\
      cam & 36.57 &	34.10 &	56.68 &	84.33 &	\textbf{47.47} &	69.12 &	87.56 \\
      can & 68.80 &	12.90 &	18.43 &	31.34 &	\textbf{72.81} &	87.56 &	98.16 \\
      cat & \textbf{41.82} & 26.27 &	37.33 &	55.30 &	24.42 &	41.01 &	61.29 \\
  driller & 63.51 &	42.86 &	57.14 &	76.04 &	\textbf{88.48} &	95.85 &	99.54 \\
     duck &	27.23 & 31.34 &	47.00 &	67.28 &	\textbf{62.67} &	78.80 &	93.09 \\
   eggbox &	69.85 & 16.59 &	22.58 &	40.09 &	\textbf{78.34} &	91.71 &	98.16 \\
     glue &	\textbf{80.02} & 11.98 &	23.04 &	32.72 &	42.40 &	57.14 &	79.72 \\
 holepunc &	42.63 & 12.90 &	20.74 &	30.88 &	\textbf{76.50} &	88.94 &	97.70 \\
     iron &	74.97 & 16.59 &	25.81 &	40.55 &	\textbf{76.50} &	91.24 &	97.70 \\
     lamp &	71.11 & 23.04 &	35.48 &	58.99  &\textbf{84.79} &	94.47 &	97.24 \\
    phone &	\textbf{74.74} & 22.12 &	29.03 &	42.86 &	64.98 &	82.03 &	96.31 \\
  \hline
  \end{tabu}%
  \caption{ADD.}
    \label{tab:Linemod_ADD}
\end{table}

\subsection{Full 6D pose estimation}
We evaluated the performance of our method for the full 6D pose estimation on T-LESS~\cite{tless}, another RGBD dataset providing 20 scenes of texture-less objects. In each scene, a few objects are placed on a board and images are acquired with the camera placed on a semi-sphere of radius \SI{75}{\cm}. Compared to LINEMOD, several objects can be used in a scene, allowing us to recover the full 6D pose of the camera. We tested our method on a scene with 6 objects and 500 images, split in two halves for training and tests. We obtained the ellipsoid of each object directly from the CAD model and generated our training images. Table~\ref{tab:TLESS_pose} shows the 6D-pose errors obtained with our method and compares it with Gaudilliere \etal \cite{gaudilliere:hal-02886633} (inscribed ellipses). This confirms the improvement made by our predicted ellipses, that was previously demonstrated on the position only.



\begin{table}[t]
  \scriptsize%
	\centering%
  \begin{tabu}{|c|c|c|}
  \hline
    
   \textbf{Method} & Gaudilliere \textit{et. al.} \cite{gaudilliere:hal-02886633} & Ours\\
   \hline
    Error in orientation (in degrees) & 3.15 ($\pm$1.96) & \textbf{2.65 ($\pm$1.40)} \\
    Error in position (in cm) & 4.09 ($\pm$2.42) & \textbf{2.23 ($\pm$2.05)} \\
  \hline
  \end{tabu}%
  \caption{Median error and standard deviation of the orientation and position estimates.}
    \label{tab:TLESS_pose}
\end{table}


\begin{figure}
    \centering
    \includegraphics[width=\linewidth]{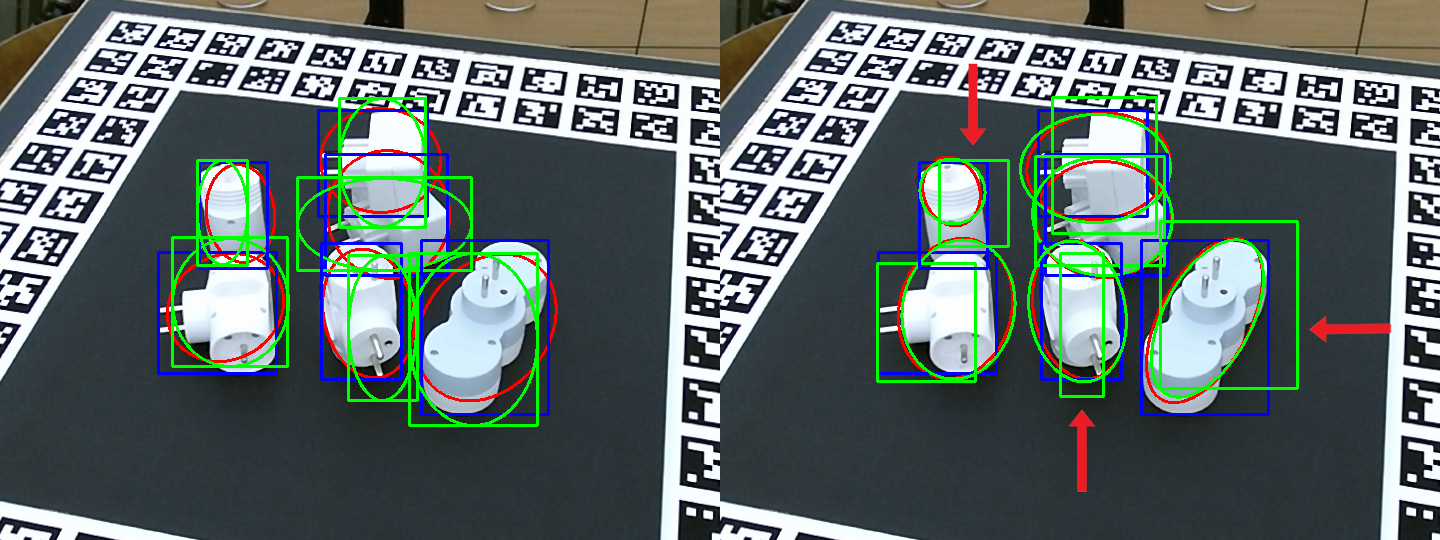}
    \caption{Inscribed ellipses (left) vs Predicted ellipses (right). Noisy BBs used for cropping and extracting the ellipses used for pose computation are in green. Ground truth projection of the ellipsoids are in red and ground truth objects BBs are in blue. Note that, despite noisy crops, the predicted ellipses fit much better to the ground truth projections.}
    \label{fig:tless_influence_of_noise_img}
\end{figure}

\subsection{Important points analysis}

\subsubsection{Robustness to detection noise}
\begin{figure}[htb]
    \centering
    \includegraphics[width=\linewidth]{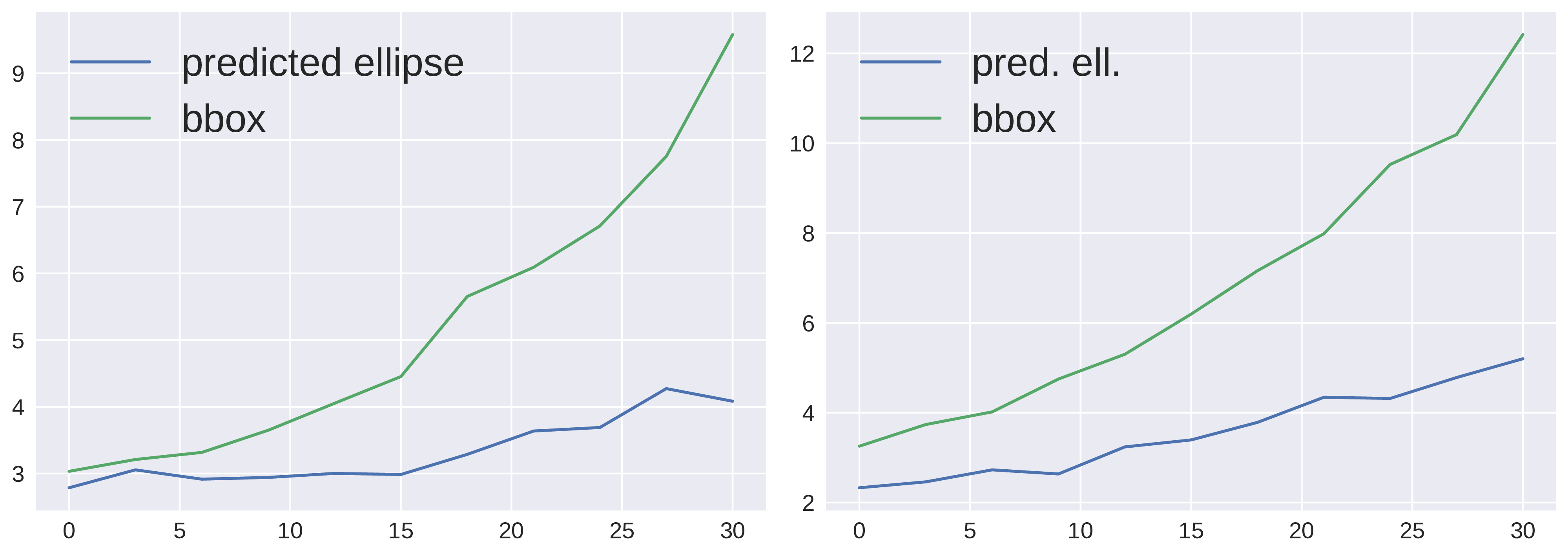}
    \caption{Influence of noisy BB detections. Left: Orientation error (in degrees). Right: Position error (in cm). Both horizontal axes represent the half-range of the noisy horiz. and vert. shifts applied on the 2 corners defining the BB.}
    \label{fig:tless_influence_of_noise}
\end{figure}

We show here that a major strength of our learning-based approach is its robustness to noisy object detections. As explained before, we use a two-step approach for detecting an object in the form of an ellipse. First, the object is detected with a 2D box, and then, this box is transformed into an oriented ellipse. The object detections step is not the focus of this work, but is nevertheless crucial. In the previous experiments, this step was completely simulated using a 2D box given by the projection of the 3D model. This is not totally realistic and, thus, we measured the influence of noisy objects detections on the pose estimation.
On the one hand, it is easy to see that spatial noise on the BB has a direct impact on its inscribed ellipse.
On the other hand, our decoupling between the box and the ellipse provides a much better robustness to variance in the detection (Fig. \ref{fig:tless_influence_of_noise_img} and \ref{fig:tless_influence_of_noise}). This is especially true for the objects marked with the arrows in Fig.~\ref{fig:tless_influence_of_noise_img}. Even though the crop passed to the prediction network does not contain the whole object, the inferred ellipses are still correct. This robustness is mostly achieved thanks to our data augmentation which randomly shifts the image (equivalent to shifting the BB before cropping).

\subsubsection{Influence of the reconstructed ellipsoid}

\begin{figure}
    \centering
    \includegraphics[width=0.8\linewidth]{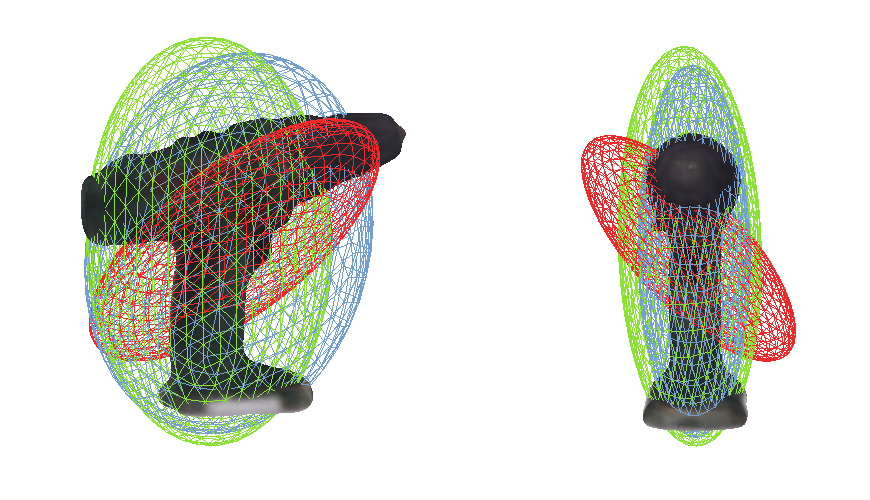}
    \caption{The three different ellipsoids used in our experiment.}
    \label{fig:Linemod_different_ellipsoids}
\end{figure}

In this experiment, we  illustrate the influence of the ellipsoid chosen as approximation of the target object. For that, we used three different ellipsoids (Fig.~\ref{fig:Linemod_different_ellipsoids}). The first one (green) was obtained by reconstruction from five images, the second one (blue) by directly fitting to the CAD model and the third one (red) was the same ellipsoid manually stretched and rotated. The results, available in Table~\ref{tab:Linemod_different_ellipsoids}, are very similar, which indicates that the choice of ellipsoid has no real influence and thus, that the network is able to learn the projection of more-or-less any ellipsoid. This is an important point of our method, because this decoupling between the ellipse and the object box detection allows us to use, for example, only a sub-part of the object.

\begin{table}[t]
  \scriptsize%
	\centering%
  \begin{tabu}{|c|c|c|c|}
  \hline
    
   \textbf{Metric} & Reprojection error & Position error & ADD\\

   \textbf{Threshold} & 5 pixels error & 5 cm & 10\% of diameter\\
   \hline
Ellipsoid 1 & \textbf{96.77} & 94.47 & \textbf{84.79} \\
Ellipsoid 2 & \textbf{96.77} & \textbf{94.93} & 84.33 \\
Ellipsoid 3 & 96.31 & 92.17 & 82.49 \\
  \hline
  \end{tabu}%
  \caption{Results obtained with different ellipsoids.}  \label{tab:Linemod_different_ellipsoids}
\end{table}

\subsubsection{Comparison with instance segmentation}
The previous experiments clearly show the superiority of our method compared to the axis-aligned ellipses. To get around this axes alignment problem, an alternative could be to fit an ellipse to a mask predicted by an instance-segmentation network.
We chose Mask R-CNN \cite{maskRCNN}, which obtains state-of-the-art results for this task and fine-tuned it on the \textit{driller}. Table~\ref{tab:Linemod_mask} shows that the ellipses fitted to the predicted masks (\textit{Ell. Mask R-CNN}) or even to the ground truth masks (\textit{Ell. GT masks}) do not reach the same level of accuracy as our predicted ellipses. This confirms our intuition presented in section  ~\ref{ssec:intuition_for_3d}.

\begin{table}
  \scriptsize%
	\centering%
  \begin{tabu}{@{\hspace{0mm}}|@{\hspace{2mm}}c@{\hspace{2mm}}|@{\hspace{2mm}}c@{\hspace{2mm}}c@{\hspace{2mm}}c@{\hspace{2mm}}|@{\hspace{2mm}}c@{\hspace{2mm}}c@{\hspace{2mm}}c@{\hspace{2mm}}c@{\hspace{2mm}}|@{\hspace{0mm}}}
  \hline

   \textbf{Method} & \multicolumn{3}{@{\hspace{2mm}}c@{\hspace{2mm}}|@{\hspace{2mm}}}{Position error} & \multicolumn{4}{c|}{ADD}  \\
   \hline
   \textbf{Thresh.} & 5 cm & 10 cm & 15 cm & 10\% & 15\% & 20\% & 25\%\\
  \hline

Ell. Mask R-CNN & 59.91 & 94.93 & 99.08 &  38.71 & 55.76 & 68.20 & 81.57 \\
Ell. GT masks & 74.65 & \textbf{100.0} & \textbf{100.0} &  47.00 & 66.36 & 82.95 & 95.39 \\
Ours & \textbf{94.47} & \textbf{100.0} & \textbf{100.0} & \textbf{84.79} & \textbf{94.01} & \textbf{98.62} & \textbf{99.54} \\
  \hline
  \end{tabu}%
  \caption{Comparison with an ellipse directly fitted to an object segmentation.}
    \label{tab:Linemod_mask}
\end{table}

\subsubsection{New scene configuration}
\label{sec:newscene}

We explain here how our method may adapt to new scene configurations, where some objects are moved between training and runtime, without the  need to retrain the system. The underlying idea is to exploit the capabilities of the  initially trained predictor to infer non  axis-aligned ellipses which are then  used to build  the ellipsoid from a few images.  By reconstructing the scene from these  ellipses  we  should  get  ellipsoids  close  to  the  initial ones and thus avoid having to train the network again.
 This strategy has been tested on LINEMOD test images of the same sequence, in which the scene has been changed and the target object rotated. An "easy" (the ape)  and a "hard" (the driller) objects were chosen in this experiment, depending  on  how much background is visible in the object crop and how much the object shape changes with a change of viewpoint. 
Table~\ref{tab:Linemod_new_scene} shows that our method still outperforms the direct BB fitting for the easy object (ape) and reaches similar results for the hard ones (driller).
Even if our approach of cropping the object allows us to be insensitive to most of the background, some of it is still visible in the crop. This shows a limit to our light 3D supervision. Having access to a 3D model (even an approximated one, but not just an ellipsoid) would enable us to be more independent to the background by generating random background during training.

\begin{table}
  \scriptsize%
	\centering%
  \begin{tabu}{|@{\hspace{2mm}}c@{\hspace{2mm}}|@{\hspace{2mm}}c@{\hspace{2mm}}c@{\hspace{2mm}}|@{\hspace{2mm}}c@{\hspace{2mm}}c@{\hspace{2mm}}|@{\hspace{2mm}}c@{\hspace{2mm}}c@{\hspace{2mm}}|}
  \hline
     \multirow{2}{*}{\textbf{Metric}} & \multicolumn{2}{@{\hspace{2mm}}c@{\hspace{2mm}}|@{\hspace{2mm}}}{Reprojection error} & \multicolumn{2}{@{\hspace{2mm}}c@{\hspace{2mm}}|@{\hspace{2mm}}}{Position error} & \multicolumn{2}{@{\hspace{2mm}}c@{\hspace{2mm}}|@{\hspace{0mm}}}{ADD}\\
     &  \multicolumn{2}{@{\hspace{2mm}}c@{\hspace{2mm}}|@{\hspace{2mm}}}{5 pixels} & \multicolumn{2}{@{\hspace{2mm}}c@{\hspace{2mm}}|@{\hspace{2mm}}}{5cm} & \multicolumn{2}{@{\hspace{2mm}}c@{\hspace{2mm}}|@{\hspace{0mm}}}{10\% of diameter}\\
    \hline
   \textbf{Ellipse} & inscr. &  pred. & inscr. & pred. & inscr. & pred. \\
      \hline
   
ape & 94.99 & \textbf{96.74} & 89.72 & \textbf{93.99} & 29.07 & \textbf{46.87} \\
driller & \textbf{21.17} & 17.15 & 47.81 & \textbf{52.19} & \textbf{25.91} & 24.45 \\

  \hline
  \end{tabu}%
  \caption{Comparison between the use of inscribed or predicted ellipses in case of a new scene configuration, without retraining.}
  \label{tab:Linemod_new_scene}
\end{table}

\subsection{Application on a real scenario}\label{sec:watch}
We demonstrate our full system on a real case using the WatchPose dataset \cite{yang:hal-02735272}, which provides pose annotated images of 10 objects in industrial environments. For each object, around 200 images are provided. Half of them were acquired "near" the object (around \SI{60}{\cm}) and the rest farther away (\SI{1.4}{\m}). This dataset differs from the previous ones and is more representative of a real application. Indeed, the images were acquired with a hand-held smartphone without specific care about motion noise, reflections and shadows. 

We follow the protocol  explained in Section~\ref{ssec:data_generation} to build the ellipsoids and generate the ellipse annotations. The BBs of the ellipses were  used to fine-tune our object detection network whereas the ellipses were used to train our ellipse prediction network. Finally, the two trained networks are used to relocalize the camera on test images. We tested two scenarios: an easy one, where a subset of \textit{near} and \textit{far} images were used for training, and a hard one, where training was done only on \textit{near} images and testing on \textit{far} images.

\begin{figure}
    \centering
    \includegraphics[width=\linewidth]{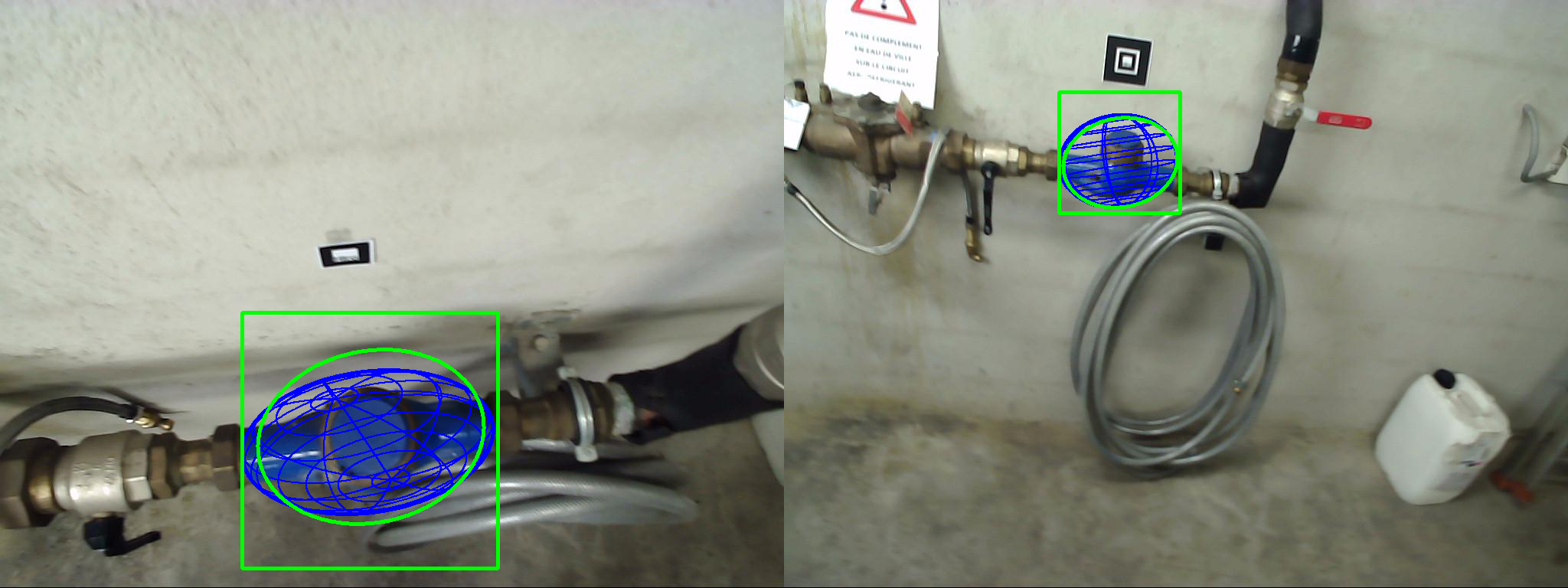}
    \caption{Predicted ellipses and ground truth projections of the ellipsoid in WatchPose.}
    \label{fig:WatchPose_results}
\end{figure}

Our experiments show that our method is able to estimate the camera position with a median error around \SI{3}{\cm} in the \textit{easy} case and \SI{6.4}{\cm} for the \textit{hard} one. We compared our approach to Gaudilliere's method \cite{gaudilliere:hal-02170784}. 
Our 3D-supervised method shows a significant improvement (see Tables~\ref{tab:WatchPose_easy},~\ref{tab:WatchPose_hard} and Fig.~\ref{fig:WatchPose_results}). For comparison with a global method (where the network uses the whole image), we trained PoseNet \cite{KendallGC15} during 5000 epochs on the \textit{near} images. While the mean position error obtained by testing also on \textit{near} images was quite good (around \SI{3}{\cm}), the generalization to more distant images was not satisfactory at all (a mean position error of \SI{80}{\cm} on \textit{far} images). This confirms that our learning on local object patches has a better generalization ability to handle new viewpoints.

\begin{table}
  \scriptsize%
	\centering%
  \begin{tabu}
{|c|c|c|c|c|c|}
  \hline
      \multirow{2}{*}{\textbf{Method}} & \textbf{Median error} & \multicolumn{4}{c|}{\textbf{Threshold}} \\
      \cline{3-6}
      
       &  \textbf{(in mm)} & 5 cm & 10 cm & 15 cm & 20 cm \\
  \hline
Gaudilliere \textit{et. al.} \cite{gaudilliere:hal-02170784} & 84.11 & 27.87 & 57.38 & 77.05 & 83.61  \\
Ours & \textbf{30.75} & \textbf{76.27} & \textbf{96.61} & \textbf{98.31} & \textbf{100}  \\
  \hline
  \end{tabu}%
  \caption{Position error in the \textit{easy} case.}
  \label{tab:WatchPose_easy}
\end{table}

\begin{table}
  \scriptsize%
	\centering%
  \begin{tabu}
{|c|c|c|c|c|c|}
  \hline
      \multirow{2}{*}{\textbf{Method}} & \textbf{Median error} & \multicolumn{4}{c|}{\textbf{Threshold}} \\
      \cline{3-6}
      
       & \textbf{(in mm)} & 5 cm & 10 cm & 15 cm & 20 cm \\
  \hline
Gaudilliere \textit{et. al.}\cite{gaudilliere:hal-02170784}& 120.31 & 14.61 & 38.20 & 66.29 & 85.39  \\
Ours & \textbf{63.90} & \textbf{34.01} & \textbf{75.00} & \textbf{90.91} & \textbf{96.59}  \\
  \hline
  \end{tabu}%
  \caption{Position error in the \textit{hard} case.}
    \label{tab:WatchPose_hard}
\end{table}

\section{Conclusion}

In this paper, we proposed a method for object-based pose estimation which does not require an accurate model of the scene. Its main component  is a new 3D-aware ellipse prediction network. By learning from different viewpoints, the network is able to map the object appearance to ellipse parameters which are coherent with the projection of the object ellipsoidal abstraction, and thus, improves the computed pose a lot. Three key aspects of the method are its robustness to variance in the box detection boundaries, its good invariance to the chosen ellipsoidal abstraction and its minimal amount of manual annotations required, making the method of large practical interest.





{\small
\bibliographystyle{ieee}
\bibliography{pose}
}

\end{document}